\documentclass[conference]{IEEEtran}
\IEEEoverridecommandlockouts

\pdfoutput=1

\usepackage[letterpaper, bottom=1in, right=0.625in, left=0.625in, top=0.75in]{geometry}
\usepackage{cite}
\usepackage{amsmath,amssymb,amsfonts,amsthm,bm}
\usepackage{graphicx}
\usepackage{textcomp}
\usepackage{xcolor}
\usepackage{caption}
\usepackage{subcaption}
\usepackage{enumitem}
\usepackage{cuted}
\usepackage{units}
\usepackage{multirow}
\usepackage[normalem]{ulem}
\usepackage[font=footnotesize,skip=1pt]{caption}
\usepackage[protrusion=true,expansion=true]{microtype}
\usepackage{balance}
\usepackage{mathtools}
\usepackage[linesnumbered,ruled,vlined]{algorithm2e}
\usepackage{algpseudocode}
\usepackage{array}

\allowdisplaybreaks

\DeclareMathOperator*{\argmax}{arg\,max}
\def\BibTeX{{\rm B\kern-.05em{\sc i\kern-.025em b}\kern-.08em
    T\kern-.1667em\lower.7ex\hbox{E}\kern-.125emX}}

\begin{document}

\title{Complexity Reduction in Machine Learning-Based Wireless Positioning: Minimum Description Features}

\author{\IEEEauthorblockN{Myeung Suk Oh\IEEEauthorrefmark{1}, Anindya Bijoy Das\IEEEauthorrefmark{1}, Taejoon Kim\IEEEauthorrefmark{2}, David J. Love\IEEEauthorrefmark{1}, and Christopher G. Brinton\IEEEauthorrefmark{1}}
\IEEEauthorblockA{\IEEEauthorrefmark{1}Electrical and Computer Engineering, Purdue University, West Lafayette, IN, USA}
\IEEEauthorblockA{\IEEEauthorrefmark{2}Electrical Engineering and Computer Science, the University of Kansas, Lawrence, KS, USA}
\IEEEauthorblockA{\IEEEauthorrefmark{1}\{oh223, das207, djlove, cgb\}@purdue.edu, \IEEEauthorrefmark{2}taejoonkim@ku.edu}}

\maketitle

\begin{abstract}
    A recent line of research has been investigating deep learning approaches to wireless positioning (WP). Although these WP algorithms have demonstrated high accuracy and robust performance against diverse channel conditions, they also have a major drawback: they require processing high-dimensional features, which can be prohibitive for mobile applications. In this work, we design a positioning neural network (P-NN) that substantially reduces the complexity of deep learning-based WP through carefully crafted minimum description features. Our feature selection is based on maximum power measurements and their temporal locations to convey information needed to conduct WP. We also develop a novel methodology for adaptively selecting the size of feature space, which optimizes over balancing the expected amount of useful information and classification capability, quantified using information-theoretic measures on the signal bin selection. Numerical results show that P-NN achieves a significant advantage in performance-complexity tradeoff over deep learning baselines that leverage the full power delay profile (PDP).
\end{abstract}


\vspace{-1mm}
\section{Introduction and Related Work}\label{sec:introduction}

Wireless positioning (WP) systems provide location awareness in many mobile applications, from intelligent vehicles to inventory control.
Typically, WP is conducted using a set of sensors that exchange signals with a target device to obtain measurements that are informative for location estimation.
Ultra-wideband (UWB) sensors are popularly used for this purpose, as they communicate on a large bandwidth that provides high distance resolution\footnote{In the IEEE 802.15.4 standard~\cite{UWB20}, UWB radio pulses are designed with the maximum duration of $2$ ns, which yields a distance resolution of $0.6$ m.} for accurate positioning.
UWB is also known to have high signal-to-noise ratio (SNR) and penetration ability, from which more reliable WP can be performed~\cite{Gezici09}.

Existing WP algorithms can be mostly categorized into two classes: geometric methods and fingerprinting methods~\cite{Mazhar17}.
With geometric methods, distance-dependent measurements, e.g., received signal strength (RSS) and time of arrival (TOA), are first acquired by the sensors.
Then, traditional estimation techniques, e.g., weighted least squares or gradient descent, are applied to predict the target location.
However, geometric methods are generally prone to high errors when the channel condition is harsh.
For example, TOA measurements under a strong non-line-of-sight (NLOS) condition are often unreliable, and compensation techniques are needed to recover performance~\cite{Guvenc09}.

Fingerprinting methods, on the other hand, take a data-driven approach, relying on a pre-acquired set of labeled measurements (i.e., with the location information being available for each measurement).
The labeled data can be used either for non-parametric estimation as new measurements arrive, e.g., through nearest-neighbor methods, or for training parametric models, e.g., support vector machines (SVM).

For fingerprinting methods, large dimensional data, e.g., power delay profile (PDP), is often used to achieve robustness against diverse channel conditions, and the models employed to handle large dimensional data are becoming increasingly complex as well.
Recently, WP based on deep learning has been considered~\cite{Fayyad20}, where neural network (NN) approaches have shown improved performance across different channel conditions and positioning environments, e.g., via convolutional neural networks (CNN)~\cite{Nguyen20} and gated recurrent units (GRU)~\cite{Nguyen21}.

Nevertheless, using PDP as learning features for WP imposes a large bandwidth and/or long latency on a sensor network, as it must be measured and stored for each positioning instance.
Also, NNs with high-dimensional input features may require high computational power and associated hardware costs to support real-time positioning rates~\cite{Fontaine20}.
These constraints can be undesirable in mobile settings where both latency and cost are critical factors, e.g., consider the requirements of WP in vehicular applications.
On the other hand, there exist some works in WP that rely on lower dimensional input features, e.g., the approach in~\cite{Zheng23} where NNs combined with a linear estimator operate on TOA and RSS measurements.
However, their performance is still heavily impacted by channel conditions, and need to be combined with additional learning tasks like ranging error detection~\cite{Kim23} anyway.
This emphasizes a steep tradeoff between performance/robustness and complexity in WP.

\textbf{Summary of contributions.} In this work, we develop a novel WP technique which reduces complexity compared with PDP-based deep learning without significantly impacting performance.
We summarize our contributions as follows:
\begin{itemize}[leftmargin=4mm]
    \item We design a positioning neural network (P-NN) that employs the largest power measurements and their temporal locations as its features. Compared to PDP, the feature set has significantly reduced dimensions yet still provides information needed to conduct accurate WP.
    \item As a component of P-NN, we develop a method for adapting the size of our features to preserve performance based on the channel conditions. Our method adopts principles of model order selection and uses the criterion which we formulate based on information-theoretic and classification capability metrics that quantify the impact of varying the number of power measurements used.
    \item We provide a set of numerical experiments to evaluate P-NN. The results show that our minimum description feature set provides accuracies that approach the PDP-based baseline while using less than 20\% of the feature size, thus achieving a desirable performance-complexity tradeoff.
\end{itemize}

\vspace{-1mm}
\section{System Model}\label{sec:system_model}
\vspace{-0.5mm}

\begin{figure}[!t]
    \centering
    \begin{subfigure}[!h]{0.5\linewidth}
        \centering
        \includegraphics[width=1\linewidth]{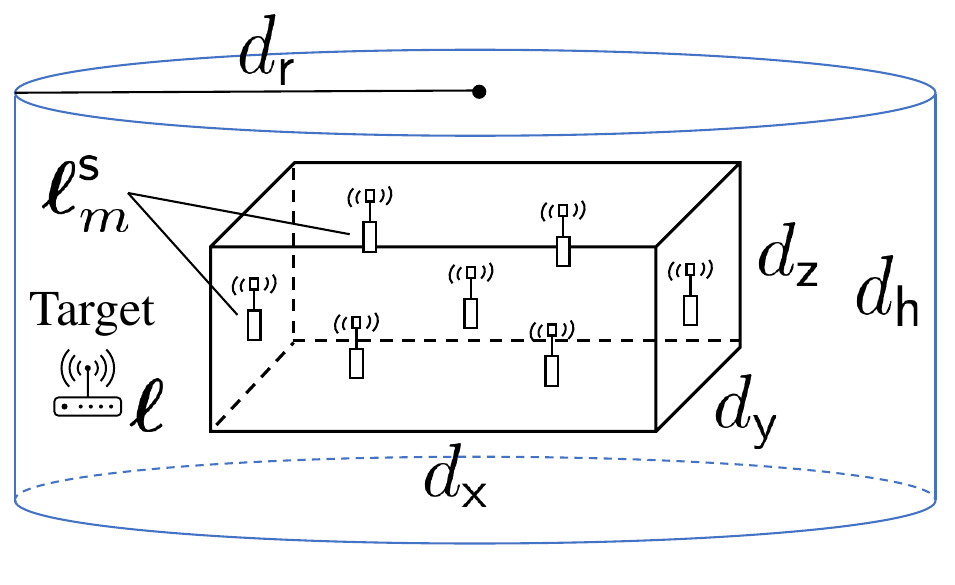}
        \caption{Layout of sensor and target spaces}
        \label{fig:system_layout}
    \end{subfigure}
    \begin{subfigure}[!h]{0.45\linewidth}
        \centering
        \includegraphics[width=0.87\linewidth]{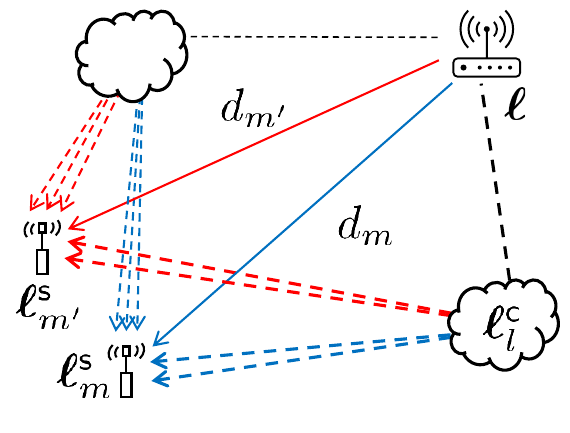}
        \caption{Channel propagation with clusters}
        \label{fig:channel}
    \end{subfigure}
    \caption{Illustrations of positioning spaces (left) and channel propagation (right).}
    \vspace{-5mm}
\end{figure}

As shown in Fig.~\ref{fig:system_layout}, we assume $M$ single-antenna sensors in a rectangular sensor space defined by the length parameters $d_\mathsf{x}$, $d_\mathsf{y}$, and $d_\mathsf{z}$.
We denote the location of sensor $m\in\{0,1,\ldots,M-1\}$ using $\boldsymbol{\ell}^\mathsf{s}_m=[x^\mathsf{s}_m, y^\mathsf{s}_m, z^\mathsf{s}_m]^\top$.
We are interested in a target that is positioned outside the sensor space but inside a cylindrical target space defined by the radius $d_\mathsf{r}$ and height $d_\mathsf{h}$.
We assume both the sensor and target spaces to be centered at $(0,0,0)$, and set $d_\mathsf{h} > d_\mathsf{z}$ and $d_\mathsf{r} > \|\frac{d_\mathsf{x}}{2}+\frac{d_\mathsf{y}}{2}\|_2$ such that the sensor space is always placed inside the target space.

\begin{figure}[!t]
    \centering
    \includegraphics[width=0.9\linewidth]{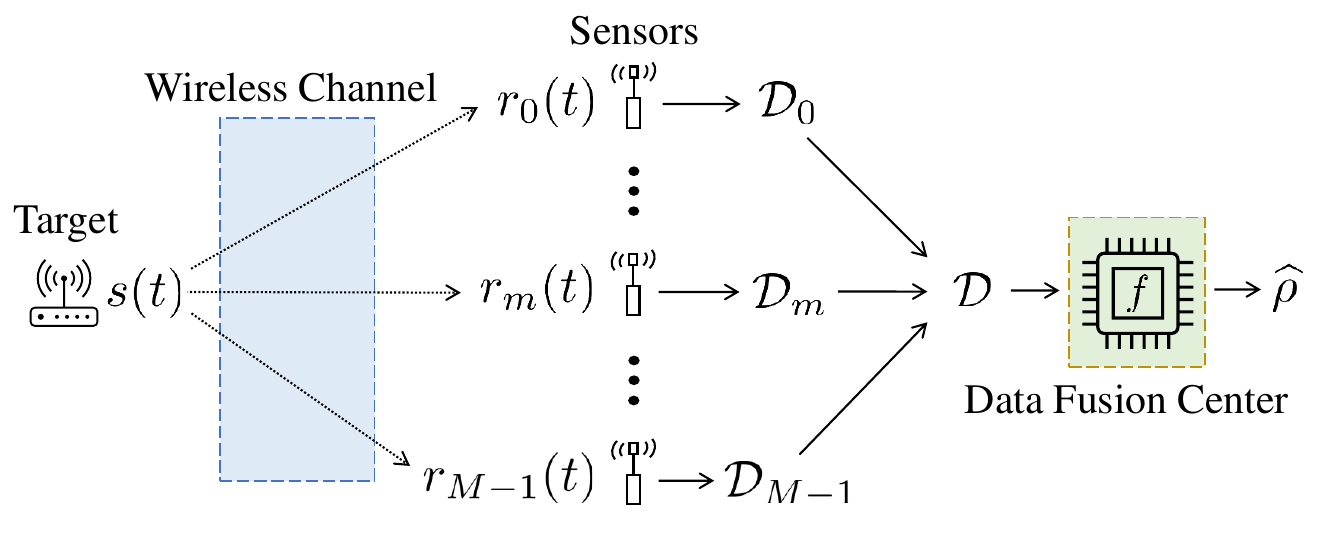}
    \caption{An overall diagram on wireless positioning.}
    \label{fig:system_model}
    \vspace{-5mm}
\end{figure}

The overall WP procedure is illustrated in Fig.~\ref{fig:system_model}.
Suppose that a target located at $\boldsymbol{\ell}=[x,y,z]^\top$ transmits a reference signal pulse $s(t)$ that is known to both the target and sensors.
The signal received by sensor $m$ is then expressed as~\cite{Molisch04}
\vspace{-1mm}
\begin{equation}
	r_{m}(t)\hspace{-0.5mm}=\hspace{-0.5mm}\sum^{L}_{l=0}\hspace{-0.5mm}\sum^{K_l-1}_{k=0}\hspace{-0.5mm}\alpha_{m,l,k}\hspace{0.5mm}s\Big(t-\frac{d_m}{c}-T_{m,l}-\tau_{m,l,k}\Big) +\,w_{m}(t),
    \label{eq:received_signal}
\end{equation}
where $L$ is the number of propagation paths imposed by channel clusters present in the target space (i.e., $l=0$ refers to the line-of-sight (LOS) path), and $K_l$ is the number of rays existing in each path $l$.
We denote the complex channel gain using $\alpha_{m,l,k}=a_{m,l,k}e^{j\phi_{m,l,k}}$, where $a_{m,l,k}$ is the weight obeying Nakagami-$\mu_{m,l,k}$ distribution with the scale value $\Omega_{m,l,k}$ and $\phi_{m,l,k}$ is the uniformly distributed phase.
$w_m(t)$ is the zero-mean complex Gaussian noise with variance $\sigma^2_{m}$.
 
With $d_{m}$ denoting the Euclidean distance between the target and sensor~$m$ (Fig.~\ref{fig:channel}) and $c$ being the speed of light, ${d_m}/{c}$ represents the TOA of the LOS path.
$T_{m,l}$ is the relative delay of path $l$ with respect to the LOS path, which is expressed~as
\vspace{-1mm}
\begin{equation}
	\hspace{-1mm}T_{m,l} =
	\begin{cases}
		0\quad\quad\quad\quad\quad\quad\quad\quad\quad\;\text{if}\;\;l=0,\text{ (LOS path)} \\
		\frac{\|\boldsymbol{\ell}^{\mathsf{c}}_l-\boldsymbol{\ell}\|_2+\|\boldsymbol{\ell}^{\mathsf{s}}_m-\boldsymbol{\ell}^{\mathsf{c}}_l\|_2-d_m}{c}\;\;\;\text{if}\;\;l > 0,
	\end{cases}
	\label{eq:cluster_delay}
    \vspace{-1mm}
\end{equation}
where $\boldsymbol{\ell}^{\mathsf{c}}_l=[x^\mathsf{c}_l, y^\mathsf{c}_l, z^\mathsf{c}_l]^\top$ is the location of cluster that imposes path $l\in\{1,\ldots,L\}$.
$\tau_{m,l,k}$ is the relative delay of ray $k$ with respect to $T_{m,l}$.
Hence, $\tau_{m,l,0}=0$, $\forall m,l$.
For $k>0$, we assume each ray follows Poisson process of the ray arrival rate $\kappa$~\cite{Molisch04}.

The pathloss of each path is expressed as~\cite{Molisch04}
\vspace{-1mm}
\begin{equation}\hspace{-1mm}
    \beta_{m,l,k}=\mathbb{E}[a_{m,l,k}^2]=\overline{P}_{\hspace{-0.5mm}m}\hspace{-0.5mm}\left({d_{m}}/ {\overline{d}_{m}}\right)^{\hspace{-1mm}-\xi} \hspace{-1mm}S^{\mathsf{s}}_{m}S^{\mathsf{c}}_{l}\, e^{-\frac{T_{m,l}}{\Gamma}-\frac{\tau_{m,l,k}}{\gamma}},\hspace{-1mm}
    \label{eq:NLOS_pathloss}
    \vspace{-1mm}
\end{equation}
where $\overline{P}_{\hspace{-0.5mm}m}$, $\overline{d}_{m}$, and $\xi$ are respectively the reference power, reference distance, and pathloss exponent.
$S^{\mathsf{s}}_{m}$ and $S^{\mathsf{c}}_{l}$ are the random shadowing applied to sensor $m$ and path $l$, respectively.
$\Gamma$ and $\gamma$ are respectively the path and ray decaying constants.
With~\eqref{eq:NLOS_pathloss}, each pathloss becomes strongly dependent on the channel propagation distance, which allows the channel paths to convey spatial correlation.

To avoid signal interference, we assume that the signal $s(t)$ is transmitted over a frame of duration $T_\mathsf{f}$ such that $ T_\mathsf{f} > \max_{m,l,k}(\frac{d_m}{c}+T_{m,l}+\tau_{m,l,k})$.
Upon receiving the frame, each sensor breaks it down to $N_\mathsf{b}=\lfloor\frac{T_\mathsf{f}}{T_\mathsf{g}}\rfloor$ temporal bins, where $T_\mathsf{g}$ is the integration period.
Then, for the signal of bandwidth $W$, the power contained in each temporal bin $n\in\{0,1,\ldots,N_\mathsf{b}-1\}$ of sensor $m$ is measured via energy detection~\cite{Urkowitz67,Dardari08}
\vspace{-1.5mm}
\begin{equation}
    \varepsilon_{m,n} = \frac{1}{2W}{\textstyle \sum}^{2WT_\mathsf{g}-1}_{i=0}\Big\vert r_m\Big(nT_\mathsf{g}+\frac{i}{2W}\Big)\Big\vert^2,
    \label{eq:bin_power}
    \vspace{-1.5mm}
\end{equation}
and $\boldsymbol{\varepsilon}_m = [\varepsilon_{m,0},\varepsilon_{m,1},\ldots,\varepsilon_{m,N_\mathsf{b}-1}]^\top$ becomes the instant PDP vector measured at sensor $m$. 

\begin{figure}[!t]
    \centering
    \begin{subfigure}[!h]{0.44\linewidth}
        \centering
        \includegraphics[width=1\linewidth]{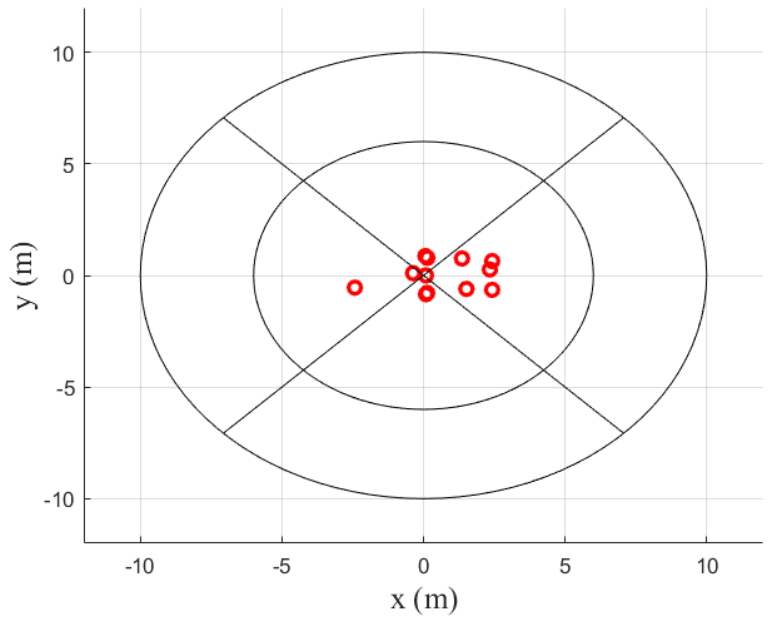}
        \label{fig:4x2}
        \vspace{-4mm}
    \end{subfigure}
    \begin{subfigure}[!h]{0.44\linewidth}
        \centering
        \includegraphics[width=1\linewidth]{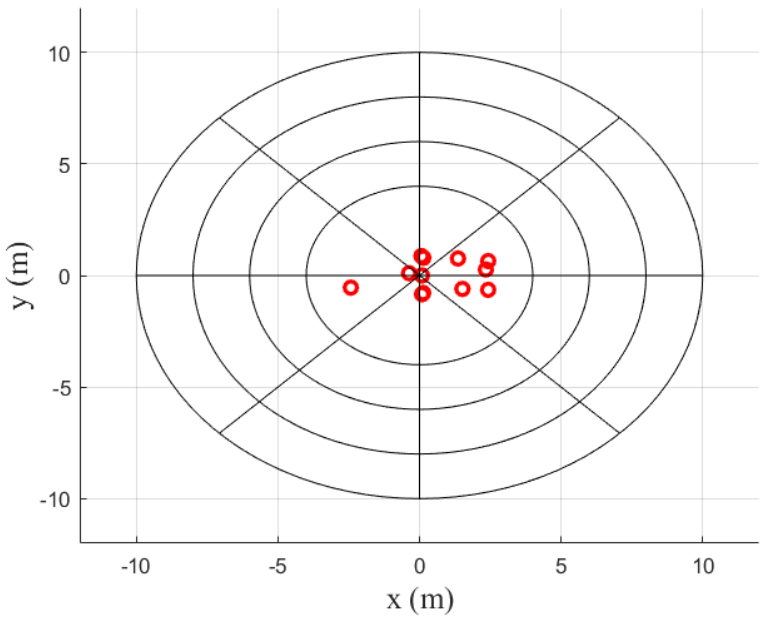}
        \label{fig:8x4}
        \vspace{-4mm}
    \end{subfigure}
    \caption{Zone layouts with $N_\mathsf{z}=8$ (left) and $N_\mathsf{z}=32$ (right). Red circles indicate sensor positions.}
    \label{fig:zone_layouts}
    \vspace{-5mm}
\end{figure}

Each sensor $m$ generates a data set $\mathcal{D}_m$ from $\boldsymbol{\varepsilon}_m$ and transfers it to the data fusion center (DFC).
Using the collected set $\mathcal{D}=\{\mathcal{D}_m\}_{m=0}^{M-1}$, the DFC estimates the target~location.
In this work, we frame our WP as an $N_\mathsf{z}$-zone classification task for the following reasons.
First, rather than coordinate-level localization, positioning via $N_\mathsf{z}$ spatial zones is often sufficient in many mobile applications.
Note that the value of $N_\mathsf{z}$ can be adjusted to satisfy the positioning sensitivity.
Second, it is more difficult to obtain coordinate-labeled training data than the zone-labeled one.
Hence, we define our positioning task using a function $f:\mathcal{D}\rightarrow\widehat{\rho}$, where $\widehat{\rho}\in\{0,1,\ldots,N_\mathsf{z}-1\}$ is the output indicating one of the $N_\mathsf{z}$ zones.
Let $\rho\in\{0,1,\ldots,N_\mathsf{z}-1\}$ denote the zone in which the target is truly located.
Then, the target is correctly positioned if $\widehat{\rho}=\rho$.
In Fig.~\ref{fig:zone_layouts}, we provide example layouts for $N_\mathsf{z}=8$ and $N_\mathsf{z}=32$, where the zones are created using radius and angle for mobile settings.

\vspace{-1mm}
\section{Positioning Neural Network (P-NN)}\label{sec:proposed_network}
\vspace{-0.5mm}

\subsection{Features of Minimum Description Length}\label{ssec:feature}
\vspace{-0.5mm}

Many WP algorithms directly use PDP data (i.e., $\mathcal{D}=\{\boldsymbol{\varepsilon}_m\}_{m=0}^{M-1}$) to achieve good performance.
Processing such high-dimensional data, however, may increase the operation requirement (e.g., bandwidth, memory, and power).
Here, we follow the principle of minimum description length (MDL)~\cite{Wax85}, which defines that the best model for describing data is one with the smallest size, and propose to use only a small number of the largest power measurements and their temporal locations.

Suppose that each sensor $m$ receives the signal $r_m(t)$ and measures the PDP vector $\boldsymbol{\varepsilon}_m$ of size $N_\mathsf{b}$.
The elements of $\boldsymbol{\varepsilon}_m$ are then sorted to yield
$\boldsymbol{\varepsilon}^{\text{ord}}_m=[\varepsilon_{m,0}^{\text{ord}},\varepsilon_{m,1}^{\text{ord}},\ldots,\varepsilon_{m,N_\mathsf{b}-1}^{\text{ord}}]^\top$, where $\varepsilon_{m,0}^{\text{ord}}\geq\varepsilon_{m,1}^{\text{ord}}\geq\ldots\geq\varepsilon_{m,N_\mathsf{b}-1}^{\text{ord}}$. Next, we denote~the~index~vector $\boldsymbol{b}^{\text{ord}}_m=[b_{m,0}^{\text{ord}},b_{m,1}^{\text{ord}},\ldots,b_{m,N_\mathsf{b}-1}^{\text{ord}}]^\top$, where $b_{m,n}^{\text{ord}}$~is~the~index of the element $\varepsilon_{m,n}^{\text{ord}}$ in $\boldsymbol{\varepsilon}_m$ (i.e., $b_{m,n}^{\text{ord}}$ indicates the temporal location in $\boldsymbol{\varepsilon}_m$ where the $n$-th largest power has been measured).
The sensor then takes the first $F$ entries of $\boldsymbol{\varepsilon}^{\text{ord}}_m$ and $\boldsymbol{b}^{\text{ord}}_m$ to generate $\mathcal{D}_m = \{\varepsilon^{\text{ord}}_{m,0},\ldots,\varepsilon^{\text{ord}}_{m,F-1},b^{\text{ord}}_{m,0},\ldots,b^{\text{ord}}_{m,F-1}\}$ of size $2F$ and transfers it to the DFC, resulting in a set $\mathcal{D}$ of size~$2FM$.

The key motivation for our feature set is an assumption that information needed for accurate WP is more likely present in the temporal bins of the largest powers.
Effective TOA estimation algorithms, e.g.,~\cite{Dardari08,Giorgetti13}, are based on this assumption and use the power threshold to detect signals.
Since both RSS and TOA of the detected signals become useful information for WP~\cite{Mazhar17}, we use both $\boldsymbol{\varepsilon}^{\text{ord}}_m$ and $\boldsymbol{b}^{\text{ord}}_m$ to generate our feature set.

Using PDP is informative as the entire $N_\mathsf{b}M$ measurements are perceived as an image for NNs to train and learn.
However, if only a small fraction of $N_\mathsf{b}$ measurements actually convey useful information, it is more desirable to process those measurements only.
However, taking the largest powers from $N_\mathsf{b}$ measurements (i.e., the first $F$ entries of $\boldsymbol{\varepsilon}^{\text{ord}}_m$) can essentially lose information within the time domain.
Hence, we directly include the temporal information (i.e., the first $F$ entries of $\boldsymbol{b}^{\text{ord}}_m$) into our feature set.

Compared to having a PDP of size $N_\mathsf{b}$, using our feature~set reduces the dimension by a factor of $\frac{2F}{N_\mathsf{b}}$ (e.g., $F=5$~and~$N_\mathsf{b}=100$ yield the size reduction by $\frac{1}{10}$).
Since deep learning algorithms (e.g., CNN of per-layer complexity that quadratically increases with feature dimensions~\cite{Vaswani17}) typically involve large data to be stored, transferred, and/or processed, reduction in feature dimensions can result in benefits such as less storage, smaller bandwidth, and lower computational complexity.

\vspace{-1mm}
\subsection{Network Architecture and Operation}\label{ssec:architecture}
\vspace{-0.5mm}

\begin{figure}[!t]
    \centering
    \includegraphics[width=0.9\linewidth]{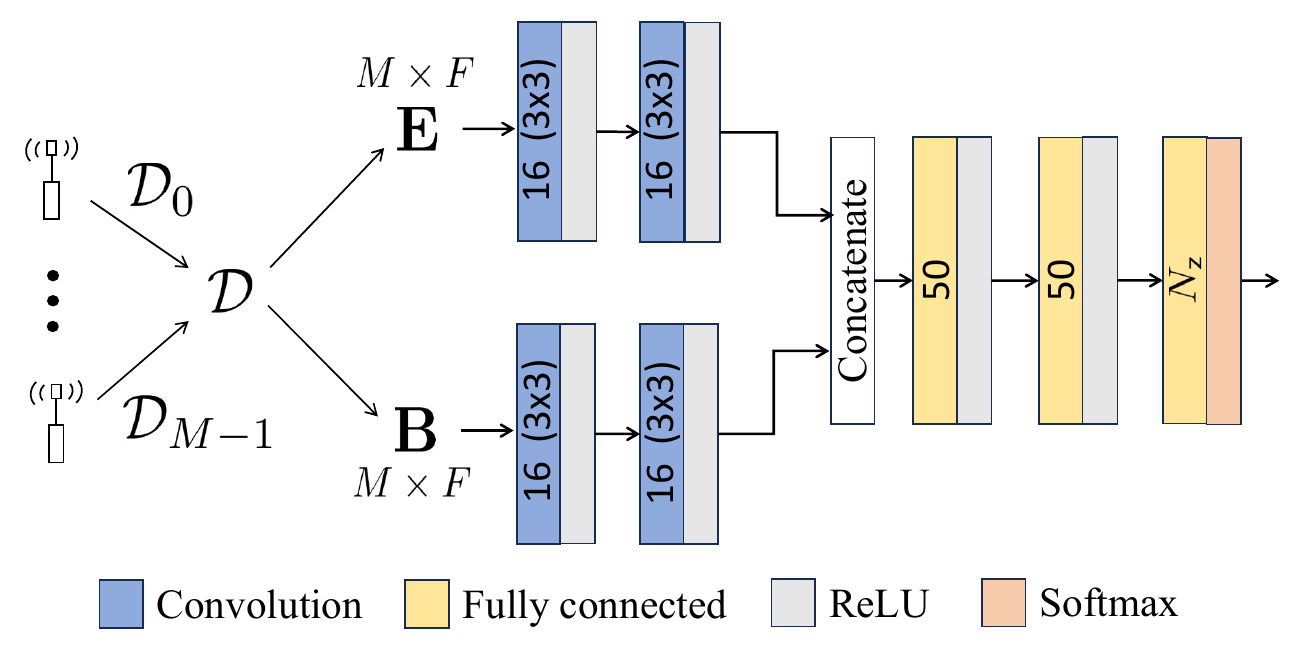}
    \caption{Architecture of our positioning neural network (P-NN).}
    \label{fig:NN_structure}
    \vspace{-5mm}
\end{figure}

The overall architecture of our P-NN is illustrated in Fig.~\ref{fig:NN_structure}.
From the collected data $\mathcal{D}$, we separate the power and time measurements, normalize them (using mean and standard deviation)~\cite{Ghasempour23}, and generate two $M\times F$ matrices $\mathbf{E}=$ $[\varepsilon^{\text{ord}}_{0,0},\ldots,\varepsilon^{\text{ord}}_{0,F-1};\ldots;\varepsilon^{\text{ord}}_{M-1,0},\ldots,\varepsilon^{\text{ord}}_{M-1,F-1}]$ and $\mathbf{B}=[b^{\text{ord}}_{0,0},$ $\ldots,b^{\text{ord}}_{0,F-1};\ldots;b^{\text{ord}}_{M-1,0},\ldots,b^{\text{ord}}_{M-1,F-1}]$.
We feed each $\mathbf{E}$ and $\mathbf{B}$ into a separate NN first to handle the data obtained from two different domains.
Here we use two convolutional layers with rectified linear unit (ReLU) activation to capture spatial correlation across both the measurements and sensors.
The outputs of two separate networks are then flattened and concatenated to be fed to a set of two fully connected (FC) layers with ReLU activation.
The last layer is designed with $N_\mathsf{z}$ neurons and softmax activation to output a classification vector that is directly translated to $\widehat{\rho}$.
The latter set of FC layers is to combine the information separately extracted from $\mathbf{E}$ and $\mathbf{B}$ and determine the output for our zone-based positioning task.

To train our P-NN, we pre-acquire a training set of size $D$, where each data point indexed by $i \in \{0,1,\ldots,D-1\}$ consists of the set $\mathcal{D}^{(i)}=\{\mathcal{D}^{(i)}_m\}_{m=0}^{M-1}$ and zone index $\rho_i$ for its label.
The network is trained offline via stochastic gradient descent (SGD).
During the testing phase, the feature set $\mathcal{D}$ is obtained from the sensors in real-time and forward-fed through the NN to determine the positioning outcome $\widehat{\rho}$.

\vspace{-1.5mm}
\section{Adaptive Feature Size Selection}\label{sec:size_selection}
\vspace{-0.5mm}

As discussed in Sec.~\ref{ssec:feature}, the $F$ largest powers and their temporal locations are collected from each sensor to form our feature set of size $2FM$.
Here we develop a strategy to adaptively select the value of $F$ as the number of measurements to be taken by each sensor for accurate WP varies by channel conditions.
To determine the value of $F$, we adopt the principle of model order selection~\cite{Akaike74} and develop a unique feature size selection method.
In the following, we define three metrics that are used to evaluate the effectiveness of our feature set.

\subsubsection{Information coming from $F$ signal bins}
Note that taking the $F$ largest power measurements for our feature set can be seen as assuming $F$ out of $N_\mathsf{b}$ bins to contain the signal.
Since each sensor measures the power according to~\eqref{eq:bin_power}, these $F$ signal-contained bins are assumed to follow non-central chi-square distribution~\cite{Urkowitz67}, which we approximate using central chi-square distribution of probability density function (PDF) given as~\cite{Giorgetti13}
\vspace{-1mm}
\begin{equation}
    f(x;\psi^2,\lambda,\nu) = \left(\frac{1}{2\eta^2}\right)^{\frac{\nu}{2}}\frac{x^{\frac{\nu}{2}-1}}{\Gamma(\frac{\nu}{2})}\exp\left(-\frac{x}{2\eta^2}\right),
\label{eq:non-central}
\vspace{-1mm}
\end{equation}
where $\eta^2=\sqrt{\frac{2\nu\psi^4+4\psi^2\lambda+(\nu\psi^2+\lambda)^2}{\nu(2+\nu)}}$ with $\psi^2$, $\lambda$, and $\nu$ being the non-central chi-square parameters and $\Gamma(\cdot)$ is the Gamma function.
The rest $N_\mathsf{b}-F$ noise-only bins are assumed to follow central chi-square distribution~\cite{Urkowitz67} of the PDF given as
\vspace{-1mm}
\begin{equation}
    f(x;\psi^2,\nu) = \left(\frac{1}{2\psi^2}\right)^{\frac{\nu}{2}}\frac{x^{\frac{\nu}{2}-1}}{\Gamma(\frac{\nu}{2})}\exp\left(-\frac{x}{2\psi^2}\right).
\label{eq:central}
\vspace{-1mm}
\end{equation}
Note that, with $\lambda=0$, only $\psi^2$ and $\nu$ characterize~\eqref{eq:central}.

Using multiple measurements of $\boldsymbol{\varepsilon}^{\text{ord}}_{m}$ from each sensor as samples, we can compute $\overline{\boldsymbol{\varepsilon}}^{\text{ord}}=[\overline{\varepsilon}_{0}^{\text{ord}},\overline{\varepsilon}_{1}^{\text{ord}},\ldots,\overline{\varepsilon}_{N_\mathsf{b}-1}^{\text{ord}}]^\top$, where $\overline{\varepsilon}_{n}^{\text{ord}}$ is the power of the $n$-th largest temporal bin averaged over both the sensors and measurements.
Using~\eqref{eq:non-central} and~\eqref{eq:central}, we define the joint PDF of $F$ non-central and $N_\mathsf{b}-F$ central chi-square variables and derive the likelihood of having $\overline{\boldsymbol{\varepsilon}}^{\text{ord}}$ as~\cite{Giorgetti13}
\vspace{-2mm}
\begin{align}
    &\hspace{-2mm}\ln f(\bar{\boldsymbol{\varepsilon}}^{\text{ord}};\psi_0^2,\ldots,\psi_{N_\mathsf{b}-1}^2,\lambda_0,\ldots,\lambda_{F-1},\nu) \nonumber \\
    &\hspace{-2mm}=\sum_{n=0}^{F-1}-\frac{\nu}{2}\ln(2\eta^2_n)+\frac{\nu-2}{2}\ln(\bar{\varepsilon}^{\text{ord}}_n)-\ln\Gamma\Big(\frac{\nu}{2}\Big)-\frac{\bar{\varepsilon}^{\text{ord}}_n}{2\eta^2_n} \nonumber \\
    &\hspace{-1.5mm}+\hspace{-1mm}\sum_{n=F}^{N_\mathsf{b}-1}\hspace{-1mm}-\frac{\nu}{2}\ln(2{\psi}^2_n)+\frac{\nu-2}{2}\ln(\bar{\varepsilon}^{\text{ord}}_n)-\ln\Gamma\Big(\frac{\nu}{2}\Big)-\frac{\bar{\varepsilon}^{\text{ord}}_n}{2{\psi}^2_n}.
\label{eq:log-likelihood}
\end{align}

\vspace{-1.5mm}
\noindent Note that~\eqref{eq:log-likelihood} is characterized by $N_\mathsf{b}$ values of $\psi^2_n$, $F$ values of $\lambda_n$, and a single value of $\nu=2WT_\mathsf{g}$.
Since we do not have the knowledge of $\{\psi^2_n\}_{n=0}^{N_\mathsf{b}-1}$ and $\{\lambda_n\}_{n=0}^{F-1}$ to evaluate~\eqref{eq:log-likelihood}, we estimate each term using
\vspace{-1mm}
\begin{equation}
    \psi^2_F = \frac{1}{N_\mathsf{b}-F}\sum_{n=F}^{N_\mathsf{b}-1}\bar{\varepsilon}^\text{ord}_n\approx\psi^2_n,\;\;\forall n=0,\ldots,N_\mathsf{b}-1,
\label{eq:psi_estimated}
\end{equation}
\begin{equation}
    \lambda^{(F)}_n = \bar{\varepsilon}^\text{ord}_n - \psi^2_F\approx\lambda_n,\;\;\forall n=0,\ldots,F-1.
\label{eq:lambda_estimated}
\vspace{-1mm}
\end{equation}
Using~\eqref{eq:psi_estimated} and~\eqref{eq:lambda_estimated}, we now define the estimated likelihood of having $\overline{\boldsymbol{\varepsilon}}^{\text{ord}}$ when the $F$ largest powers are taken for our feature set (i.e., $F$ bins are assumed to contain signals) as
\vspace{-1mm}
\begin{equation}
    \mathsf{LL}_F = \ln f(\bar{\boldsymbol{\varepsilon}}^{\text{ord}};\psi_F^2,\ldots,\psi_F^2,\lambda^{(F)}_0,\ldots,\lambda^{(F)}_{F-1},\nu).
    \label{eq:LL_F}
    \vspace{-1mm}
\end{equation}
For a given $\overline{\boldsymbol{\varepsilon}}^{\text{ord}}$, the value of~\eqref{eq:LL_F} varies by $F$, and we utilize this metric to evaluate the expected amount of information when $F$ measurements are taken for our feature set.
Note that the log-likelihood is an effective metric popularly used for the information theoretic model order selection~\cite{Akaike74,Wax85,Giorgetti13}.


\subsubsection{Information acquisition probability} Another metric we define is the probability of acquiring the useful information when we consider the $F$ largest power measurements.
Due to the time-varying nature of wireless channels, the power across the $N_\mathsf{b}$ temporal bins are randomly measured at each positioning instance.
In other words, despite the effort to generate our feature set using only the signal-contained bins, it is possible for the set to include measurements from the noise-only bins.
Such a case is not desirable since data with no useful information can degrade the performance of our P-NN.

Thus, for a given value of $F$, we quantify the chance of our feature set to take measurements from the signal-contained bins.
Recall that taking the $F$ largest power measurements is to assume $F$ signal-contained bins out of $N_\mathsf{b}$.
First, we define $P_\mathsf{th}^{(F)}=(\overline{\varepsilon}_{F-1}^{\text{ord}}+\overline{\varepsilon}_{F}^{\text{ord}})/2$ be the power threshold that separates the first $F$ bins from the rest $N_\mathsf{b}-F$ bins. 
Our logic is that the feature set will likely include these signal-contained bins if their power is measured greater than $P_\mathsf{th}^{(F)}$.
Hence, using~\eqref{eq:psi_estimated} and~\eqref{eq:lambda_estimated}, we define the probability of a signal-contained bin $n\in\{0,\ldots,F-1\}$ to have the power greater than $P_\mathsf{th}^{(F)}$ as~\cite{Dardari08}
\begin{align}
    p^{(F)}_{n} & = \mathbb{P}\left\{\frac{\varepsilon^{\text{ord}}_n}{\psi^2_{F}}>\frac{P_\mathsf{th}^{(F)}}{\psi^2_{F}}\;\bigg\vert\;\frac{\lambda^{(F)}_n}{\psi^2_{F}}\right\} \nonumber \\
    & = Q_{\frac{\nu}{2}}\left(\sqrt{2(\lambda_n^{(F)}/\psi^2_{F})^2},\sqrt{2P_\mathsf{th}^{(F)}/\psi^2_{F}}\right),
\label{eq:high_signal}
\end{align}

\vspace{-1mm}
\noindent where $Q_{\frac{\nu}{2}}\left(\cdot,\cdot\right)$ is the $\frac{\nu}{2}$-th order Marcum Q-function~\cite{Marcum60}.
Based on~\eqref{eq:high_signal}, we define the acquisition probability of our $F$ largest powers to include the measurements from $f\in\{0,1,\ldots,F\}$ signal-contained bins as
\vspace{-1mm}
\begin{equation}
    \textstyle\mathsf{P}^{(F)}_f=\sum_{\boldsymbol{q}\in\mathcal{Q}^{(F)}_f}\prod_{i=1}^{F}(p^{(F)}_{i-1})^{\boldsymbol{q}[i]}(1-p^{(F)}_{i-1})^{(1-\boldsymbol{q}[i])},
\label{eq:safe_capture}
\vspace{-2mm}
\end{equation}
where $\mathcal{Q}^{(F)}_f$ is the set of all $F$-length binary vectors containing $f$ ones (i.e., $\mathcal{Q}^{(F)}_f$ considers all $\frac{F!}{f!(F-f)!}$ cases where $f$ out of $F$ bins have their power greater than $P_\mathsf{th}^{(F)}$).
The product term in~\eqref{eq:safe_capture} computes the joint probability of each case in $\mathcal{Q}^{(F)}_f$, and the summation provides the overall probability.
Note that~\eqref{eq:safe_capture} quantifies the chance of taking $f$ useful measurements when we consider the $F$ largest measurements for our feature set.
 
\subsubsection{Inter-zone Kullback-Leibler divergence}

Dissimilarity among the class distributions is one of the key factors that impact classification performance, and how we form our feature set directly affects this dissimilarity.
Hence, for a given value of $F$, we propose to quantify the dissimilarity across the data samples from each zone via Kullback-Leibler (KL) divergence and use it for our feature size selection.
To evaluate KL divergence, the PDFs must be known.
Since we only have empirical measurements (i.e., training data), we take the k-nearest neighbors (KNN) density estimation approach to directly estimate the KL divergence~\cite{Perez08}.
If we subgroup the training data by each zone in terms of our feature set and denote each group using $\mathcal{D}^{\mathsf{z}}_z$ for $z\in\{0,1,\ldots,N_\mathsf{z}-1\}$, the estimated KL divergence between the zone $z$ and $z'$ using the KNN density estimation with $u$ nearest neighbors is given by
\vspace{-1mm}
\begin{equation}
    \hspace{-1mm}\widehat{D}_u(P_z||P_{z'}) = \frac{F}{|\mathcal{D}^{\mathsf{z}}_z|}\hspace{-1mm}\sum_{x\in\mathcal{D}^{\mathsf{z}}_z}\hspace{-1mm}\log \frac{r_{u,z'}(x)}{r_{u,z}(x)}+\log\frac{|\mathcal{D}^{\mathsf{z}}_{z'}|}{|\mathcal{D}^{\mathsf{z}}_z|-1},
    \label{eq:KL_estimated}
    \vspace{-2mm}
\end{equation}
where $r_{u,z}(x)$ is the Euclidean distance between $x$ and its $u$-th nearest neighbor in $\mathcal{D}^{\mathsf{z}}_z$.
Now we define the average KL divergence upon taking the $F$ largest power measurements as
\vspace{-1mm}
\begin{equation}
    \textstyle
    \mathsf{KL}_F = \frac{1}{N_\mathsf{z}^2\sqrt{F}}\sum_{i=0}^{N_\mathsf{z}}\sum_{j=0}^{N_\mathsf{z}}\widehat{D}_u(P_i||P_j),
    \label{eq:KL_mean}
    \vspace{-1mm}
\end{equation}
which we use to quantify how effectively our feature set of size $2FM$ can separate the classes.
Note that, regardless of the distributions being compared,~\eqref{eq:KL_estimated} yields a steady increase with $F$ due to the volume expression used in the KNN density estimation.
Hence, a factor of $\sqrt{F}$ is applied in~\eqref{eq:KL_mean} to account for the increase in the expected Euclidean distance across $F$.

Using the metrics~\eqref{eq:LL_F},~\eqref{eq:safe_capture}, and~\eqref{eq:KL_mean}, we now formulate our feature size selection criterion, which is expressed as
\vspace{-2mm}
\begin{equation}
    F^\star \hspace{-0.5mm} = \hspace{-2mm}\argmax_{F\in[F_\mathsf{min},F_\mathsf{max}]} \hspace{-1mm}\bigg(\epsilon\underbrace{\sum_{f=0}^{F}\mathsf{P}^{(F)}_f\frac{f}{F}\overline{\mathsf{LL}_F\hspace{-0.5mm}-\hspace{-0.5mm}\mathsf{LL}_0}}_{(a)} + (1-\epsilon)\underbrace{\overline{\mathsf{KL}_F}}_{(b)}\bigg)
\label{eq:size_selection}
\vspace{-3mm}
\end{equation}
where $\overline{(\cdot)}$ implies the normalization with respect to $\max_F(\cdot)$ and $\epsilon\in[0,1]$ is the weight parameter.
Since our criterion is the weighted sum of $(a)$ and $(b)$, we force the range of both $(a)$ and $(b)$ to be $[0,1]$ by normalizing $\{\mathsf{LL}_F\hspace{-0.5mm}-\hspace{-0.5mm}\mathsf{LL}_0\}_{F=F_\mathsf{min}}^{F_\mathsf{max}}$ and $\{\mathsf{KL}_F\}_{F=F_\mathsf{min}}^{F_\mathsf{max}}$.
Our selection criterion in~\eqref{eq:size_selection} reflects two factors: the effective amount of information, i.e., $(a)$, and classification capability, i.e., $(b)$, from taking the $F$ largest powers and their temporal locations.
Note that we compute $\mathsf{LL}_F\hspace{-0.5mm}-\hspace{-0.5mm}\mathsf{LL}_0$ to exactly quantify the increase in information upon taking the $F$ largest measurements.
To account for the chance that our $F$ measurements include only $f$ ones that are actually useful, we multiply $\frac{f}{F}$ and the acquisition probability $\mathsf{P}^{(F)}_f$ to $\overline{\mathsf{LL}_F\hspace{-0.5mm}-\hspace{-0.5mm}\mathsf{LL}_0}$.

\textbf{Example.} We provide a numerical example of our feature size selection using the setting of 15dB SNR and LOS condition.
For brevity, we set $N_\mathsf{b} = 10$, $[F_\mathsf{min},F_\mathsf{max}]=[3,8]$, and~ $\nu=2$.
From the given setting, we obtain $\bar{\boldsymbol{\varepsilon}}^\text{ord}=[53.9,26.8,17.4,$ $12.5,9.46,6.35,5.22,4.06,3.76,2.55]\hspace{-0.5mm}\times\hspace{-0.5mm}10^{-7}$, where the first five entries contain the signal.
In Table~\ref{tb:example}, we provide some of the numerical values computed for the given example with $\epsilon = 0.5$.
Using the last two rows of Table~\ref{tb:example}, we evaluate our criterion values for $F\in[3,8]$ to be $\{0.79,0.76,0.89,0.88,0.87,0.85\}$ and determine $F^\star=5$ based on~\eqref{eq:size_selection}.
As shown in Table~\ref{tb:example}, the given $\bar{\boldsymbol{\varepsilon}}^\text{ord}$ provides a steady increase in $\mathsf{LL}_F\hspace{-0.5mm}-\hspace{-0.5mm}\mathsf{LL}_0$ from $F=4$ to $F=6$.
However, a larger $F$ also increases the chance of taking measurements from the noise-only temporal bins, which contributes to the decrease in $p_{n}^{(F)}$ and results in a negligible increase in the effective amount of information.
Hence, our selection criterion determines $F^\star=5$ to be the number of measurements to be taken for our features.

\begin{table}[!t]
\centering
\setlength\extrarowheight{2pt}
\setlength\tabcolsep{3pt}
\caption{Numerical values of the key parameters used in our feature size selection steps. $\psi^2_F$, $\lambda^{(F)}_n$, and $P_\mathsf{th}^{(F)}$ are in the unit of $10^{-7}$.}
\label{tb:example}
\begin{tabular}{|c|c|c|c|} 
    \hline
    $F$ & $4$ & $5$ & $6$ \tabularnewline
    \hline
    $\psi^2_F$ & $5.23$ & $4.39$ & $3.89$ \tabularnewline
    \hline
    $\{\lambda^{(F)}_n\}_{n=2}^{F-1}$ & $12.13,7.27$ & $12.97,8.11,5.07$ & $13.46,8.60,5.56,2.45$ \tabularnewline
    \hline
    $\mathsf{LL}_F\hspace{-0.5mm}-\hspace{-0.5mm}\mathsf{LL}_0$ & $4.651$ & $5.099$ & $5.326$ \tabularnewline
    \hline
    $P_\mathsf{th}^{(F)}$ & $10.98$ & $7.91$ & $5.79$ \tabularnewline
    \hline
    $\{p^{(F)}_{n}\}_{n=2}^{F-1}$ & $0.92,0.57$ & $0.99,0.83,0.52$ & $0.99,0.95,0.72,0.35$ \tabularnewline
    \hline
    $\{\mathsf{P}^{(F)}_f\}_{f=3}^{F}$ & $0.44,0.53$ & $0.09,0.49,0.42$ & $0.01,0.20,0.55,0.24$ \tabularnewline
    \hline
    $(a)$ in~\eqref{eq:size_selection} & $0.7213$ & $0.7857$ & $0.7912$ \tabularnewline
    \hline
    $\mathsf{KL}_F$ & $16.48$ & $16.64$ & $16.28$ \tabularnewline
    \hline
\end{tabular}
\vspace{-5mm}
\end{table}

\vspace{-1.5mm}
\section{Numerical Evaluation}\label{sec:numerical}
\vspace{-1mm}

We conduct a set of numerical experiments to evaluate our P-NN.
We consider a rectangular sensor space of $d_\mathsf{x}=6$~m, $d_\mathsf{y}=3$~m, and $d_\mathsf{z}=2$~m with $M=12$ sensors and a cylindrical target space of $d_\mathsf{r}=10$~m and $d_\mathsf{h}=4$~m.
We consider the residential model for UWB channel~\cite{Molisch04}, for which we generate $L$ randomly located channel clusters using Poisson distribution of mean $\overline{L}=3$ and set $K_l=6$ for all $l$.
For Nakagami distributions, we assume $\mu_{m,l,k}$ follows log-normal distribution of mean $0.67$~dB and variance $0.28$~dB and $\Omega_{m,l,k}=\beta_{m,l,k}$, $\forall m,l,k$~\cite{Molisch04}.
We set $\kappa=1.5$~ns, $\Gamma=25$~ns, $\gamma=5$~ns, and $\xi=2$ and consider $\overline{P}_m=-45$~dBm and $\overline{d}_m=1$~m for all sensors~\cite{Molisch04}.
We assume both $S^{\mathsf{s}}_{m}$ and $S^{\mathsf{c}}_{l}$ follow zero-mean log-normal distribution with 3~dB variance~\cite{Molisch04}.
We assume $W=2$ GHz, $T_\mathsf{f}=200$ ns, and $T_\mathsf{g}=2$ ns to have $N_\mathsf{b}=100$.
For each sensor $m$, we define SNR as ${\mathbb{E}[\beta_{m,0,0}}]/{\sigma^2_{m}}$, where the expectation is over the target space.
For NLOS channel, we set $\alpha_{m,0,k}=0$ for all $m$ and $k$ to remove the LOS path.~For~the~KL

\noindent divergence estimation, we use $u=30$.

\begin{figure}[!t]
    \centering
    \begin{subfigure}[!h]{0.49\linewidth}
        \centering
        \includegraphics[width=1\linewidth]{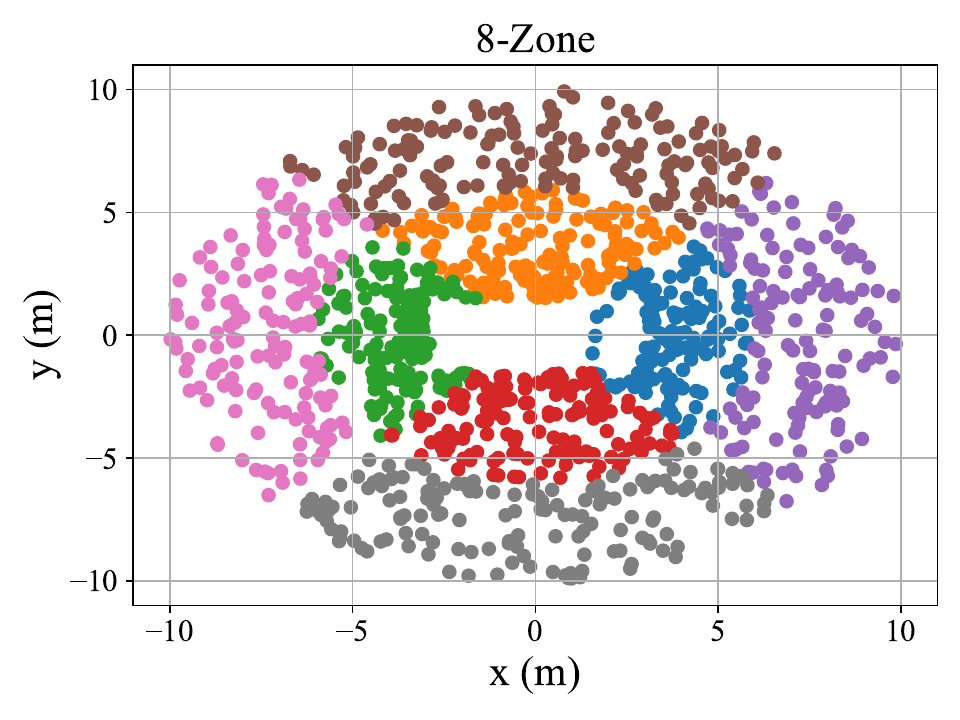}
    \end{subfigure}
    \begin{subfigure}[!h]{0.49\linewidth}
        \centering
        \includegraphics[width=1\linewidth]{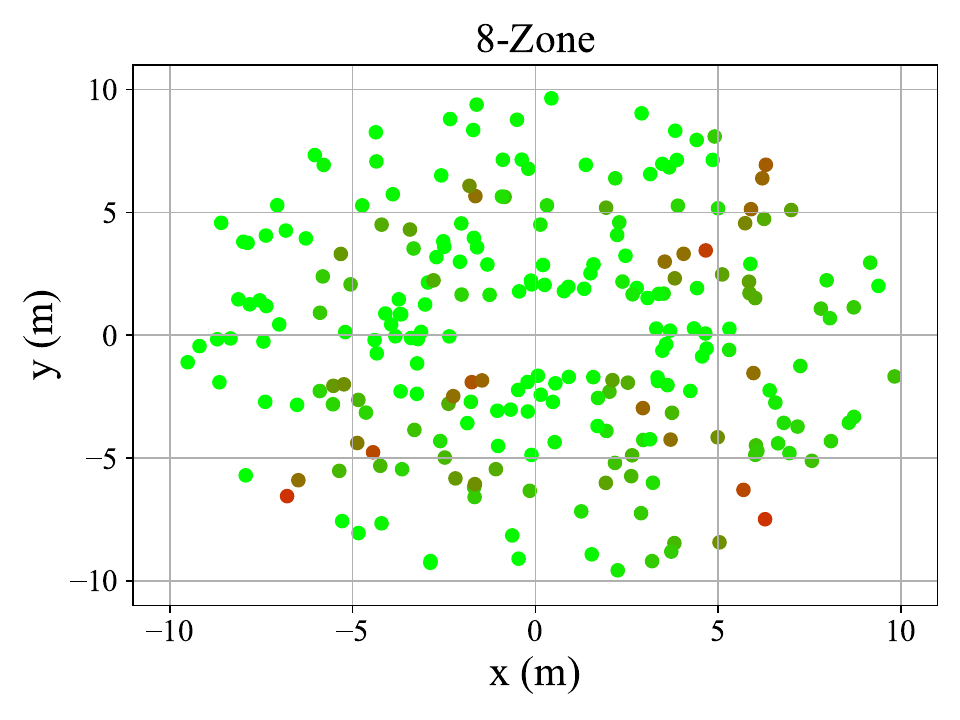}
    \end{subfigure}
    \caption{An illustration of training (left) and testing (right) sets in a 2D plane. For the training set, same color implies the same classification zone. For the testing set, redder color indicates lower classification accuracy.}
    \label{fig:sets}
    \vspace{-2mm}
\end{figure}

To compare our P-NN with existing algorithms, we consider CNN-LE~\cite{Nguyen20} and NN-LCS~\cite{Zheng23} as the baseline.
Note that CNN-LE and NN-LCS respectively use PDP and TOA/RSS as their features.
For the training phase, $D=30,000$ target locations were randomly generated, and a pair of $\mathcal{D}^{(i)}$ and $\rho_i$ was obtained for each location.
We used Adam optimizer of learning rate $0.001$, and the training was performed over $50$ epochs with the random batch size $256$.
For the testing phase, a set of $\mathcal{D}$ and $\rho$ pairs were generated from $6,000$ random target locations, and the performance of each WP algorithm was measured by comparing each predicted output $\widehat{\rho}$ with $\rho$.
An visual illustration of our training and testing sets is provided in Fig.~\ref{fig:sets}.
For statistical significance, the result was averaged over $20$ simulation runs, and five independent scenarios were used. 

\textbf{Feature size selection:} We demonstrate the effectiveness of our feature size selection method given in Sec.~\ref{sec:size_selection}.
In Table~\ref{tb:feature_size}, we provide the performance (in zone classification rate) of our P-NN using different values of $F$ over various channel conditions.
For each row, the numerical value in bold indicates the performance obtained using $F^\star$ from our method.
We observe that training our P-NN with $F^\star$ can maintain high performance with a relatively lower feature size.
This verifies that taking the largest power and time measurements constitutes minimum description features for navigating the performance-complexity tradeoff.
Overall, our feature size selection can adaptively determine the dimensions of our features and lead to high WP performance.

\begin{table}[!t]
\centering
\setlength\extrarowheight{1pt}
\setlength\tabcolsep{2.5pt}
\caption{Zone classification rates (in percent) of P-NN with different values of $F$. The rates achieved using $F^\star$ in~\eqref{eq:size_selection} are indicated in bold. We set $\epsilon=0.8 (\text{or }0.6)$ for the LOS (or NLOS) channel scenarios.}
\label{tb:feature_size}
\begin{tabular}{|c|c|c|c|c|c|c|c|c|c|} 
 \hline
  Scenario \# & SNR & $F$ = $4$ & $F$ = $5$ & $F$ = $6$ & $F$ = $7$ & $F$ = $8$ & $F$ = $9$ & $F$=$10$ \tabularnewline
 \hline
  LOS \#3 & \multirow{4}{*}{15dB} & 91.21 & 91.59 & 92.07 & 92.35 & 92.51 & 92.67 & \bf{92.82} \tabularnewline
 \cline{1-1}\cline{3-9}
  LOS \#4 & & 88.21 & 89.42 & 90.11 & 90.51 & \bf{90.88} & 90.84 & 90.89 \tabularnewline
 \cline{1-1}\cline{3-9}
 NLOS \#3 & & 76.31 & 77.25 & 77.79 & \textbf{77.80} & 78.14 & 78.25 & 78.41 \tabularnewline
 \cline{1-1}\cline{3-9}
 NLOS \#4 & & 69.67 & 72.30 & 74.48 & 75.59 & 76.00 & 76.79 & \bf{77.24} \tabularnewline
 \hline
 LOS \#3 & \multirow{4}{*}{5dB} & 68.48 & 69.67 & 70.32 & 70.71 & 71.03 & 71.14 & \bf{71.24} \tabularnewline
 \cline{1-1}\cline{3-9}
  LOS \#4 & & 69.71 & 70.50 & 70.92 & 71.45 & 72.09 & 72.24 & \textbf{72.37} \tabularnewline
 \cline{1-1}\cline{3-9}
 NLOS \#3 & & 44.19 & 44.64 & 44.94 & 45.23 & 45.12 & \bf{45.39} & 45.57 \tabularnewline
 \cline{1-1}\cline{3-9}
 NLOS \#4 & & 49.22 & 49.26 & 49.46 & \textbf{49.80} & 50.00 & 50.21 & 50.15 \tabularnewline
 \hline
\end{tabular}
\vspace{-6mm}
\end{table}

\begin{figure*}[!t]
    \centering
    \minipage{0.32\textwidth}
        \includegraphics[width=\linewidth]{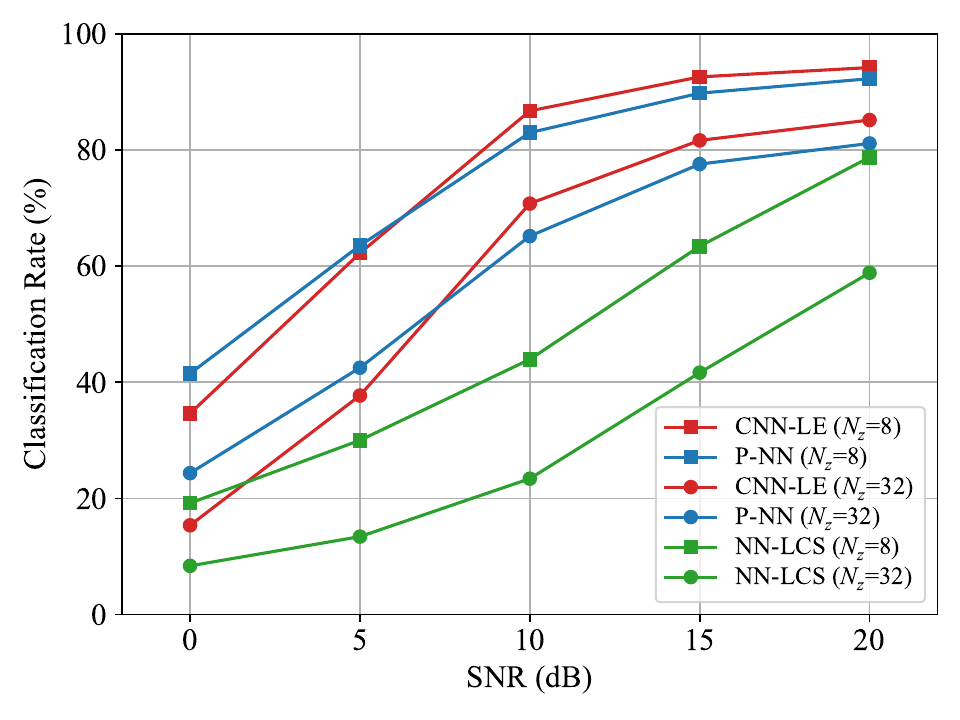}
        \caption{Performance vs. SNR of different WP algorithms with LOS channels. Feature sizes for CNN-LE and NN-LCS are $1200$ and $24$, respectively. Feature size for the proposed ranges from $72$ to $240$.}
        \label{fig:LOS_rate}
    \endminipage
    \hfill
    \minipage{0.32\textwidth} 
        \includegraphics[width=\linewidth]{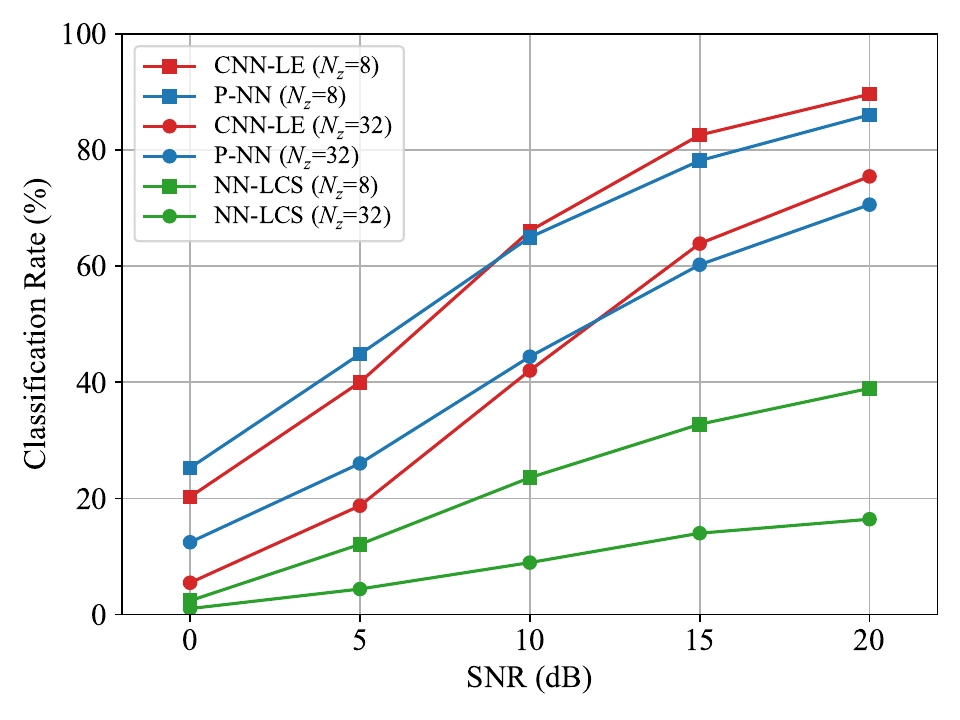}
        \caption{Performance vs. SNR of different WP algorithms with NLOS channels. Feature sizes for CNN-LE and NN-LCS are $1200$ and $24$, respectively. Feature size for the proposed ranges from $72$ to $240$.}
        \label{fig:NLOS_rate}
    \endminipage
    \hfill
    \minipage{0.32\textwidth}
        \includegraphics[width=\linewidth]{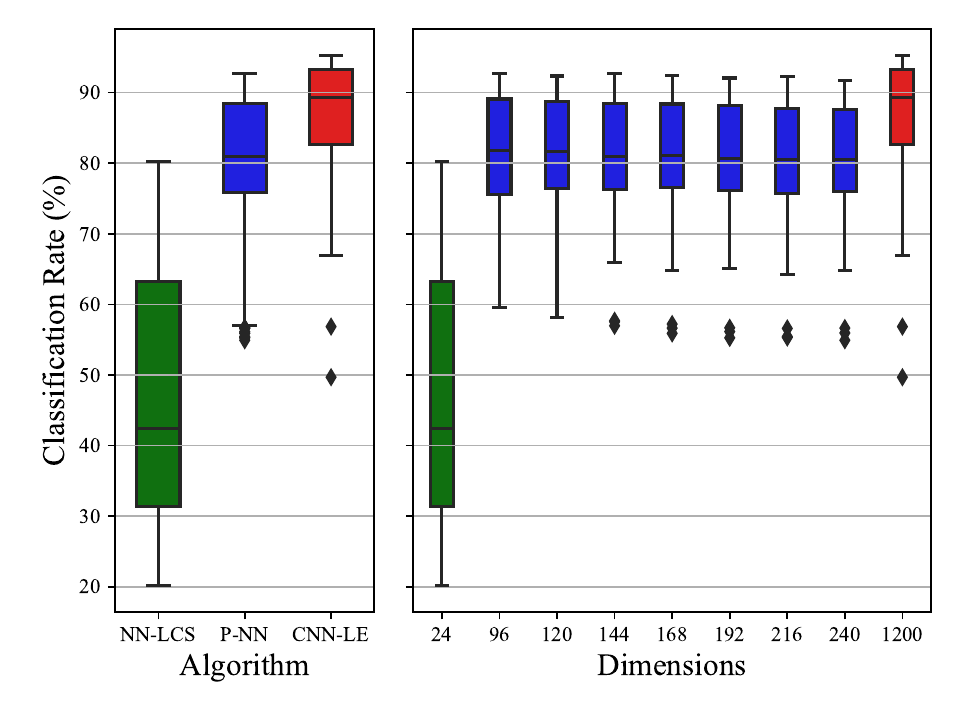}
        \caption{Classification rates obtained with $10$, $15$, and $20$ dB SNRs by different WP algorithms (left) and the number of dimensions (right). For P-NN, we consider $F\in[4,10]$.}
        \label{fig:tradeoff}
    \endminipage
\vspace{-5mm}
\end{figure*} 

\textbf{Classification performance:} Next, we compare the performance of P-NN with the baselines. 
In Figs.~\ref{fig:LOS_rate} and~\ref{fig:NLOS_rate}, we provide classification rate vs. SNR plots for LOS and NLOS channels, respectively.
For P-NN, we determine $F^\star$ from a range $[4,10]$.
We observe that the performance of NN-LCS in both plots is significantly lower, demonstrating the difficulty of achieving good WP performance from a small-sized feature.
Compared to NN-LCS, both CNN-LE and P-NN provide better performance.
Especially in low SNR, P-NN outperforms CNN-LE as it discards the measurements from noise-only bins, the power of which become greater with low SNR, and thus prevents them from being used in the NN training.
Given that the performance is competitive between CNN-LE and P-NN (i.e., one outperforms the other depending on the SNR level), our P-NN, which takes only the largest measurements from PDP, takes an advantage in the performance-complexity tradeoff.  

\textbf{Performance-complexity tradeoff:} To demonstrate the advantage of our P-NN in the performance-complexity tradeoff, we provide box plots showing the range of classification rates obtained by different WP algorithms and the number of feature dimensions in Fig.~\ref{fig:tradeoff}.
We observe that NN-LCS has the lowest dimension, but the performance range is low and has a high variance.
CNN-LE exhibits steady and high classification rate, but such a performance is achieved at the cost of utilizing high-dimensional feature.
P-NN using our feature set shows the performance similar to the one of CNN-LE at relatively low feature dimensions.
This result shows that our feature set can provide WP performance that is much more complexity-efficient.

\vspace{-1mm}
\section{Conclusions}\label{sec:conclusion}
\vspace{-0.2mm}

We have considered a WP scenario for mobile applications and proposed P-NN that utilizes a low-dimensional feature.
Our minimum description feature set takes a number of largest power measurements and their temporal positions.
For robust performance against varying channel conditions, we have proposed a method of adaptively selecting the feature size by considering the log-likelihood, acquisition probability, and KL divergence.
Numerical results have shown that using our feature set achieves positioning performance competitive to the one from using PDP and has a great performance-complexity tradeoff as compared to the baseline algorithms.
Potential directions of future works include (i) theoretical optimization on the weight parameter for the feature size selection and (ii) the extension of P-NN to consider coordinates estimation.
\vspace{-1.2mm}

\section*{Acknowledgment}
\vspace{-1mm}
This work was supported in part by the National Science Foundation (NSF) under grants EEC1941529, CNS2146171, CNS2212565, CNS2225577, and CNS2225578.
\vspace{-1mm}

\balance{
\bibliographystyle{IEEEtran}
\bibliography{IEEEfull,mybib}

\begin{thebibliography}{10}
\providecommand{\url}[1]{#1}
\csname url@samestyle\endcsname
\providecommand{\newblock}{\relax}
\providecommand{\bibinfo}[2]{#2}
\providecommand{\BIBentrySTDinterwordspacing}{\spaceskip=0pt\relax}
\providecommand{\BIBentryALTinterwordstretchfactor}{4}
\providecommand{\BIBentryALTinterwordspacing}{\spaceskip=\fontdimen2\font plus
\BIBentryALTinterwordstretchfactor\fontdimen3\font minus
  \fontdimen4\font\relax}
\providecommand{\BIBforeignlanguage}[2]{{%
\expandafter\ifx\csname l@#1\endcsname\relax
\typeout{** WARNING: IEEEtran.bst: No hyphenation pattern has been}%
\typeout{** loaded for the language `#1'. Using the pattern for}%
\typeout{** the default language instead.}%
\else
\language=\csname l@#1\endcsname
\fi
#2}}
\providecommand{\BIBdecl}{\relax}
\BIBdecl

\bibitem{UWB20}
``{IEEE} standard for low-rate wireless networks--amendment 1: Enhanced ultra
  wideband ({UWB}) physical layers ({PHYs}) and associated ranging
  techniques,'' \emph{IEEE Std 802.15.4z-2020 (Amendment to IEEE Std
  802.15.4-2020)}, pp. 1--174, 2020.

\bibitem{Gezici09}
S.~Gezici and H.~V. Poor, ``Position estimation via ultra-wide-band signals,''
  \emph{Proc. IEEE}, vol.~97, no.~2, pp. 386--403, 2009.

\bibitem{Mazhar17}
F.~Mazhar, M.~Khan, and B.~Sällberg, ``Precise indoor positioning using {UWB}:
  A review of methods, algorithms and implementations,'' \emph{Wireless Pers.
  Commun.}, vol.~97, 12 2017.

\bibitem{Guvenc09}
I.~Guvenc and C.-C. Chong, ``A survey on {TOA} based wireless localization and
  {NLOS} mitigation techniques,'' \emph{IEEE Commun. Surv. \& Tut.}, vol.~11,
  no.~3, pp. 107--124, 2009.

\bibitem{Fayyad20}
J.~Fayyad, M.~A. Jaradat, D.~Gruyer, and H.~Najjaran, ``Deep learning sensor
  fusion for autonomous vehicle perception and localization: A review,''
  \emph{Sensors}, vol.~20, no.~15, 2020.

\bibitem{Nguyen20}
D.~T.~A. Nguyen, H.-G. Lee, E.-R. Jeong, H.~L. Lee, and J.~Joung, ``Deep
  learning-based localization for {UWB} systems,'' \emph{Electronics}, vol.~9,
  no.~10, 2020.

\bibitem{Nguyen21}
D.~T.~A. Nguyen, J.~Joung, and X.~Kang, ``Deep gated recurrent unit-based {3D}
  localization for {UWB} systems,'' \emph{IEEE Access}, vol.~9, pp.
  68\,798--68\,813, 2021.

\bibitem{Fontaine20}
J.~Fontaine, M.~Ridolfi, B.~Van~Herbruggen, A.~Shahid, and E.~De~Poorter,
  ``Edge inference for {UWB} ranging error correction using autoencoders,''
  \emph{IEEE Access}, vol.~8, pp. 139\,143--139\,155, 2020.

\bibitem{Zheng23}
Z.~Zheng, S.~Yan, L.~Sun, H.~Shu, and X.~Zhou, ``{NN-LCS}: Neural network and
  linear coordinate solver fusion method for {UWB} localization in car keyless
  entry system,'' \emph{Sensors}, vol.~23, no.~5, 2023.

\bibitem{Kim23}
D.-H. Kim, A.~Farhad, and J.-Y. Pyun, ``{UWB} positioning system based on
  {LSTM} classification with mitigated {NLOS} effects,'' \emph{IEEE Internet of
  Things J.}, vol.~10, no.~2, pp. 1822--1835, 2023.

\bibitem{Molisch04}
A.~F. Molisch, K.~Balakrishnan, D.~Cassioli, C.-C. Chong, S.~Emami, A.~Fort,
  Johan, Karedal, J.~Kunisch, H.~G. Schantz, U.~G. Schuster, and K.~Siwiak,
  ``{IEEE} 802.15.4a channel model-final report,'' 2004.

\bibitem{Urkowitz67}
H.~Urkowitz, ``Energy detection of unknown deterministic signals,'' \emph{Proc.
  IEEE}, vol.~55, no.~4, pp. 523--531, 1967.

\bibitem{Dardari08}
D.~Dardari, C.-C. Chong, and M.~Win, ``Threshold-based time-of-arrival
  estimators in {UWB} dense multipath channels,'' \emph{IEEE Trans. Commun.},
  vol.~56, no.~8, pp. 1366--1378, 2008.

\bibitem{Wax85}
M.~Wax and T.~Kailath, ``Detection of signals by information theoretic
  criteria,'' \emph{IEEE Trans. Acoust., Speech, Signal Process.}, vol.~33,
  no.~2, pp. 387--392, 1985.

\bibitem{Giorgetti13}
A.~Giorgetti and M.~Chiani, ``Time-of-arrival estimation based on information
  theoretic criteria,'' \emph{IEEE Trans. Signal Process.}, vol.~61, no.~8, pp.
  1869--1879, 2013.

\bibitem{Vaswani17}
A.~Vaswani, N.~Shazeer, N.~Parmar, J.~Uszkoreit, L.~Jones, A.~N. Gomez,
  L.~Kaiser, and I.~Polosukhin, ``Attention is all you need,'' in \emph{Adv.
  Neural Inf. Process. Syst.}, vol.~30, 2017.

\bibitem{Ghasempour23}
A.~Ghasempour and M.~Martínez-Ramón, ``Electric load forecasting using
  multiple output gaussian processes and multiple kernel learning,'' in
  \emph{IEEE Symp. Ind. Electron. \& Appl. (ISIEA)}, 2023, pp. 1--6.

\bibitem{Akaike74}
H.~Akaike, ``A new look at the statistical model identification,'' \emph{IEEE
  Trans. Autom. Control}, vol.~19, no.~6, pp. 716--723, 1974.

\bibitem{Marcum60}
J.~Marcum, ``A statistical theory of target detection by pulsed radar,''
  \emph{IRE Trans. Inf. Theory}, vol.~6, no.~2, pp. 59--267, 1960.

\bibitem{Perez08}
F.~Perez-Cruz, ``{Kullback-Leibler} divergence estimation of continuous
  distributions,'' in \emph{IEEE Int. Symp. Inf. Theory}, 2008, pp. 1666--1670.

\end{thebibliography}
}

\end{document}